\documentclass[letterpaper, 10 pt, conference]{ieeeconf}  

\IEEEoverridecommandlockouts                              
\usepackage[utf8]{inputenc}
\usepackage[T1]{fontenc}
\usepackage{graphicx}
\usepackage{url}
\usepackage{hyperref}

\title{\LARGE \bf
Designing Robots to Help Women*
}

\author{Martin Cooney$^{1}$, Lena Klasén$^{2}$, and Fernando Alonso-Fernandez$^{1}$
\thanks{*We gratefully acknowledge support from the Swedish Innovation Agency (VINNOVA) for the project "AI-Powered Crime Scene Analysis"}
\thanks{$^{1}$M. Cooney and F. Alonso-Fernandez are with the School of Information Technology, Halmstad University
301 18 Halmstad, Sweden
        {\tt\small martin.daniel.cooney@gmail.com, fernando.alonso-fernandez@hh.se}}%
\thanks{$^{2}$L. Klasén is with the Computer Vision Laboratory (CVL), Department of Electrical Engineering (ISY), Linköping University, 581 83 Linköping, Sweden and Department of National Operations, Swedish Police Authority, 102 66 Stockholm, Sweden
        {\tt\small lena.klasen@liu.se}}%
}

\begin{document}

\maketitle
\thispagestyle{empty}
\pagestyle{empty}

\begin{abstract}

Robots are being designed to help people in an increasing variety of settings--but seemingly little attention has been given so far to the specific needs of women, who represent roughly half of the world's population but are underrepresented in robotics.
Here we used a speculative prototyping approach to explore this expansive design space:
First, we identified some challenges that disproportionately affect women in relation to crime, health, and daily activities, as well as opportunities for designers, which were visualized in five sketches.
Then, one of the sketched scenarios was further explored by developing a prototype, of a drone equipped with computer vision to detect hidden cameras that could be used to spy on women. 
While object detection introduced some errors, hidden cameras were identified with a reasonable accuracy of 80\% (Intersection over Union (IoU) score: 0.40).
Our aim is that these results could help spark discussion and inspire designers, toward realizing a safer, more inclusive future through responsible use of technology.
\end{abstract}

\section{INTRODUCTION}
\label{section:intro}

\begin{figure}
\centering
\includegraphics[width=.5\textwidth]{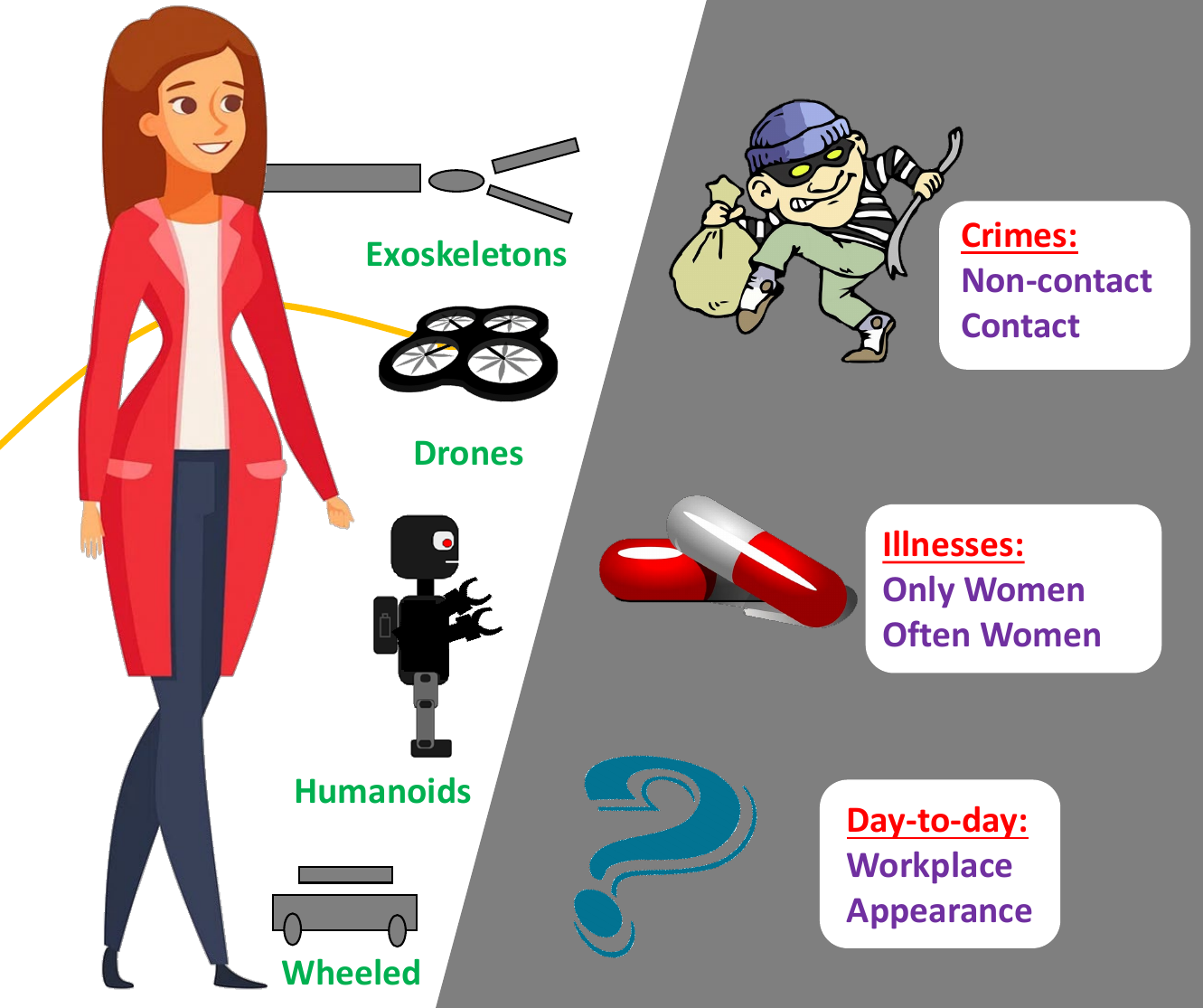}
\caption{Basic concept: robots could be designed to help women with challenges related to crime, health, and daily activities} \label{fig_intro}
\end{figure}

Within the area of feminist Human-Robot Interaction (HRI), the current paper explores how robots could be designed to help women to deal with various common challenges, as pictured in Fig.~\ref{fig_intro}.

Discrepancies can sometimes be observed between how we would like the world to function, and how the world actually functions:
We believe that people should be treated equally, with similar rights and opportunities, and that women are an important group to consider, in line with the concept of gender mainstreaming and the UN's Sustainable Development Goal 5.\footnote{\href{eige.europa.eu/gender-mainstreaming/what-is-gender-mainstreaming}{eige.europa.eu/gender-mainstreaming/what-is-gender-mainstreaming}}$^,$\footnote{\href{unwomen.org/en/node/36060}{unwomen.org/en/node/36060}}
Yet despite constituting roughly half of the human population, women have been historically marginalized, underrepresented, ignored, and restricted~\cite{lewis2020difficult}.\footnote{\href{womansday.com/life/real-women/a55991/no-women-allowed}{womansday.com/life/real-women/a55991/no-women-allowed}} 
Some beliefs can seem humorous in retrospect, like that women might not be able to ride trains as their uteruses might fly out of their bodies due to high speeds.
Other more sobering examples question rights that might seem fundamental, like the rights to vote, run, or defend, which have been granted only recently in some countries (e.g., the right to vote was granted in Liechtenstein only in 1984, the right to run with men in the Boston Marathon in the United States (US) only in 1972, and the right to carry out some combat jobs in the US army only in 2016).

As we look toward the future, in which technologies like Artificial Intelligence (AI) and robotics are expected to help people to live better lives, we find similar indications of potential marginalization:
For example, reported examples of gender bias in AI include Google preferentially showing high-paying ads to men, Google Translate defaulting to male pronouns, Amazon's hiring system preferring applications from men, and LinkedIn suggesting that female names are mistaken~\cite{tannenbaum2019sex}.\footnote{\href{https://www.reuters.com/article/idUSKCN1MK0AG/}{https://www.reuters.com/article/idUSKCN1MK0AG/}}
\footnote{\href{https://qz.com/775597/linkedins-lnkd-search-algorithm-apparently-favored-men-until-this-week}{https://qz.com/775597/linkedins-lnkd-search-algorithm-apparently-favored-men-until-this-week}}
As another example, Autonomous Vehicles (AVs) are poised to contribute to safe, efficient transport, but even basic automotive features such as seat belts, airbags, and crash dummies have up until recently seemingly not taken female body sizes into account, leading to an increased rate of injury in traffic accidents (47\% higher between 1998 and 2008 in the US); as well, cars' speech recognition systems have had trouble understanding female voices~\cite{howard2018ugly}.
Furthermore, as Winkle et al. describe, female opinions have been highly underrepresented in HRI studies, as with Human-Computer Interaction~\cite{winkle202315}: robot designers are generally not female, and examples of tackling women's needs appear to be scarce.
(Note: In using the term "women" here, we do not focus on non-binary or gender-fluid cases, although our ideation might also be relevant to some degree for such groups; as well, we follow the typical convention that "sex" refers to biological state and "gender" to social identity.)

One barrier is that the designated design space is expansive:
Women face various challenges that can seem "wicked", complex, unclear, interwoven, and difficult to solve; robots also can have various capabilities, forms (e.g., flying, wheeled, humanoid, or exoskeletal), and deployment configurations (e.g., carried, located nearby, or sent from a police station or hospital), etc.
Thus, a rough understanding of the "big picture" would seem useful for designers and policy makers to select meaningful challenges to tackle, consider potential solutions, and make informed decisions. Here, we follow a critical \emph{speculative prototyping} approach intended to elicit new insights into both theory and practice.

The remainder of the paper is structured as follows: In Section~\ref{section:review}, we briefly frame and motivate our work in comparison to previous work. 
Section~\ref{section:method} describes how we identified challenges and opportunities. 
Section~\ref{section:prototype} goes deeper into one specific scenario via a prototype, and Section~\ref{section:discussion} discusses results.

In summary, the current paper's contribution is two-fold:
\begin{itemize}
\item{{\bf Theoretical}. We identify and explore some new challenges and opportunities related to useful tasks for robot designers to help a large user group (women).}
\item{{\bf Practical}. We explore one scenario in more depth, reporting on a new proof-of-concept for a robot.}
\end{itemize}

\section{RELATED WORK}
\label{section:review}


Previous work related to women and HRI seems to have mostly focused on topics outside of the scope of the current paper, such as differences in how women and men perceive robots, or sex robots. For example, women were observed to place more trust than men in the idea of a security robot~\cite{gallimore2019trusting}, and sex robots, rather than promoting objectification of women, could actually foster liberalization and sex-positivity~\cite{kubes2019new}).

However, a few studies have explored how robot designers could address challenges specific to women.
One work that is highly relevant to the current topic is by Winkle et al., who proposed feminist HRI--highlighting the importance of examining and challenging power, in considering diverse perspectives, emotions, embodiments, and "low power" users, to achieve responsible design~\cite{winkle2023feminist}.
As well, Winkle and Mulvihill explored how robots could be used to abuse women in a domestic violence (DV) scenario, toward mitigating risks~\cite{winkle2024anticipating}.
The authors comment briefly on possibilities of using ambient sensor data or mobile apps to help detect DV, yet advise caution--referring to a study by Cookson et al. that discusses how digital interventions can be overly hyped and under-deliver, have unintended consequences and hidden costs, and alone are no "magical bullet"~\cite{cookson2023fit}.
A difference with the current work is that these studies did not seek to speculate about what kinds of tasks future robots could one day perform to help women. For example, the presence of robots with sensors could also one day help in a DV scenario, by complicating perpetrators' efforts to isolate victims and avoid having crimes recorded or observed.

Much work has also looked at how robots could use sensing to help people in general.
For example, we previously explored how a robot could seek to defend a person who is being attacked~\cite{cooney2023broad}. 
We also experimented with combining object detection and thermal trace detection to infer activities, which inspired our prototype in the current paper~\cite{cooney2017pastvision+}.
However, these works did not focus on the unique requirements of women.
Thus, it seemed useful to further explore the question of how robots could help women.

\section{SPECULATION}
\label{section:method}

To provide both theoretical and practical insights, a speculative prototyping approach was followed.
Speculative prototyping aims to "drag" potentially important future scenarios from the foggy realm of imagination closer to the real world, visualizing in a thought-provoking manner, as a means of embarking on a process of critical exploration~\cite{tironi2018speculative}.
As a first step, rapid ideation sessions were conducted among the authors, who contained a female member and had some experience with robotics, AI, computer vision, criminology, and health technology.
For both challenges and opportunities, ideas were recorded without judgement, then afterwards merged, refined by surveying related literature, and reworked with priority attributed to those that seemed valuable, different, and requiring further study.
In doing so, we aimed to roughly align ourselves with Suvin's concept of "cognitive estrangement", described by Bartolotta in terms of seeking to capture "fictional world-affordances imbued with cognitive potential"~\cite{bartolotta2022beyond}.
We also adopted a \emph{zemiological} perspective~\cite{winkle2024anticipating}; i.e., a broad view that encompasses various potential harms, in relation to three topics we identified: crime, health, and daily activities.

\subsection{Challenges Faced by Women}
\label{section:challenges}

\subsubsection{Crimes}
\label{section:crimes}

Crimes that disproportionately affect women can occur globally or locally, and at distance or in proximity.
In general, crimes that don't require close contact include voyeurism, sexting without permission, stalking, and indecent exposure; crimes that require contact include DV (including intimate partner abuse), rape (including date, acquaintance, marital, and gang rapes), molestation (grabbing, groping), child abuse and grooming, forced prostitution (including sex trafficking, sexual slavery, forced marriage, revenge porn, and forced participation in the production of pornography), and other forms of sexual assault, harassment, bullying, and violence~\cite{centers2010national}.\footnote{\href{https://pcar.org/about-sexual-violence/adults}{https://pcar.org/about-sexual-violence/adults}}
In some regions of Asia and Africa, women can also face risks of female genital mutilation, honor killings, stonings, acid throwing, blinding, persecution due to alleged witchcraft, war rape, and forced suicide (e.g., \emph{Sati}, in which a widow is forced to die on her husband's funeral pyre\footnote{\href{https://en.wikipedia.org/wiki/Sati\_(practice)}{https://en.wikipedia.org/wiki/Sati\_(practice)}}).

Various complexities exist: For example, some of the above terms overlap, and even non-sexual violence can relate to sex, since women who are smaller can be seen as easier targets~\cite{madriz2023nothing}. 
As well, crimes can be further detailed in terms of individual actions; e.g., physical violence can involve slapping, pushing, hitting, kicking, hair-pulling, choking, burning, or assault with objects such as knives or guns.

Yet despite these complexities, crimes would be important to mitigate since they affect many women.
For example, a report from 2002 indicated that two thirds of Swedish women have been sexually harassed or experienced violence~\cite{lundgren2002captured}: 
Over half of women have been sexually harassed, and around half have experienced violence (a quarter physical violence, one in three sexual violence), including around a quarter of young women 18-24 just in the past year; one out of every five women has also been threatened by a man.
Numbers are also disturbing in the US, where around two thirds of women have experienced violence, and around one fifth of women have been raped, usually by someone they knew~\cite{centers2010national}.

From the large pool of crimes described, three crimes were selected to explore in greater detail, one remote, and two requiring contact.
From the remote category, voyeurism, also related to \emph{scopophilia}, upskirting and peeping, seemed useful to explore, since it might be relatively easy to begin with, and is an important, widespread problem:
The cost of a failed intervention could be lower than for more serious crimes, given that victims often don't know they have been victimized and thus don't suffer trauma~\cite{green2020criminalizing}.
Moreover, current legislature regarding filming without consent seems to support technological approaches to limit video voyeurism, and it seems relatively clear how voyeurism can be stopped, by preventing a criminal from viewing and filming victims.
Furthermore, although actual numbers are difficult to predict, voyeurism has been described as an epidemic in Korea, with over 6000 cases reported each year~\cite{teshome2019spy}.
When footage is spread, the result can be "social death", causing some women to attempt suicide.
For such reasons, in 2018, approximately 20,000 women marched to protest spy cameras, with 200,000 signing a petition, which resulted in a government plan to hire 8000 workers to tackle the problem. 
As well, in Sweden, in an anonymous survey of 2,450 randomly selected 18–60 year-olds conducted in 1996, 7.7\% of respondents described deriving sexual satisfaction from spying on others having sex~\cite{laangstrom2006exhibitionistic}.
Thus, it could be useful for women if voyeurism could be more easily detected and prevented.

Of crimes that require proximity, two kinds that seem more difficult to tackle, but highly important, include DV and rape. 
For example, in Sweden, there are typically no witnesses to DV, as the attacks occur "behind closed doors"~\cite{pratt2014don}. As such, various statistics on DV can be found, but the real numbers, and hence the true extent of this problem, are unknown. 
As well, hospital records are often generated as a result of attacks, but police are usually not allowed to access these data; victims can also feel retraumatized and uncared for when interacting with healthcare staff.

Furthermore, recent discourse has highlighted difficulties in Sweden related to immigration and integration~\cite{pulkkinen2024swedish}:
Especially victims coming from other cultures can find themselves in a position of weakness--segregated, and with no one to talk to: 
They might not be capable of communicating in the local language, allowed to go outside, or aware of Swedish norms (i.e., no one might explicitly tell them that is not okay for a man to beat a woman).
They might also visit religious centers that reinforce non-Swedish views, like that they must obey the men in their families, or that they could be punished or killed for having a local boyfriend.
For example, views in Africa on DV can be startlingly different, with acceptance of wife-beating at 77\% in Mali and Uganda; overall, "51\% of African women report that being beaten by their husbands is justified if they either go out without permission, neglect the children, argue back, refuse to have sex, or burn the food."\footnote{\href{https://blogs.worldbank.org/en/africacan/domestic-violence-and-poverty-in-africa-when-the-husbands-beating-stick-is-like-butter}{https://blogs.worldbank.org/en/africacan/domestic-violence-and-poverty-in-africa-when-the-husbands-beating-stick-is-like-butter}}
Another key problem with DV is that repeat offenses are common over the long term~\cite{amaral2023deterrence}. For example, officers might pick up a criminal on Friday for beating a woman, then release him on Saturday, in a pattern that repeats itself each week.
The perpetrator as well can escalate, beating harder and harder each time, which can end in murder.
Some mechanism is needed to break such damaging spirals.

Various suggestions have been made: 
For example, phone calls to DV helplines can be masked, and women freed from family phone plans on request.\footnote{\href{https://ksltv.com/602266/fcc-adopts-new-cellphone-rules-designed-to-keep-domestic-violence-victims-safer}{https://ksltv.com/602266/fcc-adopts-new-cellphone-rules-designed-to-keep-domestic-violence-victims-safer}}
As well, help can be sought by drawing a black dot on one's hand, selecting a red pen at a hospital, or uttering keywords like "Angela" or "Minotaur" at a bar.\footnote{\href{www.bbc.com/news/blogs-trending-34326137}{www.bbc.com/news/blogs-trending-34326137}}$^,$\footnote{\href{themighty.com/2020/01/domestic-violence-preventionsign-
red-marker}{themighty.com/2020/01/domestic-violence-preventionsign-
red-marker}}$^,$\footnote{\href{canadianwomen.org/signal-for-help}{canadianwomen.org/signal-for-help}}
However, it's not clear how effective such existing strategies are:
Backlash effects could exist; if an aggressor learns about such a communication, the victim could be beaten more.
Furthermore, perpetrators typically seek to "gaslight" or manipulate victims into doubting themselves, by establishing narratives to maintain power and control--e.g., telling victims that they are crazy, useless, or to blame, and that this is how things must be~\cite{hailes2023they}. 
As a result of repeatedly experiencing traumatic and uncontrollable events, including physical and psychological abuse, victims often have little self-esteem left, and frequently develop posttraumatic stress disorder (PTSD), major depressive disorder (MDD), or "learned helplessness"--which contribute to submissiveness and reduce a woman's belief that her actions could lead to a positive result~\cite{bargai2007posttraumatic}.
Thus, while successful arrests could help to reduce repeated offences by approximately half~\cite{amaral2023deterrence}, as with other crimes such as sex trafficking or forced marriage, battered women often feel trapped, and it's hard for women themselves to break free.

Then, what might be needed?
As above, one challenge is getting information to the victims, so they are no longer alone.
Also, women who are victims need to be protected; something should be done without them having to start the process.

Another important problem to combat is rape, which shares similarities with DV:
Rapes are usually conducted by perpetrators known to the victim; the crime is committed in close contact; its repercussions are among the most serious of crimes (leading to chronic health problems such as PTSD, MDD, and substance abuse); help-seeking victims often experience a "second rape" in being repeatedly interrogated, doubted, blamed, warned, and discouraged; and a chronic failure to report or investigate rape has been noted~\cite{campbell2008psychological}~\cite{bryden1996rape}.
In some countries, like the US, clearance rates for rape also appear to be low (e.g., in the 60 percent range, compared to Japan in the high 90s, with rapes being about 100 times more common)~\cite{parker2001japanese}.

Challenges include legislation and processing of evidence.
For example, the European Parliament, in its recent "Artificial Intelligence Act", has banned real-time detection, on the basis of the potential for mistakes.\footnote{\href{europarl.europa.eu/news/en/press-room/20240308IPR19015/artificial-intelligence-act-meps-adopt-landmark-law}{europarl.europa.eu/news/en/press-room/20240308IPR19015/artificial-intelligence-act-meps-adopt-landmark-law}} 
More concretely, in an urban environment with surveillance cameras, it is not permitted to detect rape in real time and alert police.
Rape must happen first--everyone is required to wait--then images can be used, after judicial approval is obtained.
As such, currently much of the potential use of AI is lost, and the integrity of victims is not upheld.
While it's clear that current recognition systems are imperfect, various examples of working real-time systems exist, not just in pacemakers and brakes, but also employing computer vision, from Tesla cars doing real-time inference from cameras, to real-time pixelization of faces on television.
Furthermore, a basis exists for how to handle imperfect recognition in standards such as SOTIF (ISO/PAS 21448).\footnote{\href{https://www.iso.org/standard/77490.html}{https://www.iso.org/standard/77490.html}} 
Thus, here the challenge seems to be not merely technical but also political; a prototyping example of how robots could detect and prevent crime in real-time at an early stage could also be useful as a way to affect thought regarding legislature.

Another problem involves bottlenecks in processing evidence.
Disturbingly, it seems there are hundreds of thousands of untested "rape kits" in the US, some of which are not processed even after 30 years~\cite{lion2017bringing}.\footnote{\href{https://usafacts.org/articles/how-many-rape-kits-are-awaiting-testing-in-the-us-see-the-data-by-state}{https://usafacts.org/articles/how-many-rape-kits-are-awaiting-testing-in-the-us-see-the-data-by-state}}
This forensic evidence, which is costly and requires hours of invasive handling of victims, is sometimes not sent onward for analysis or cannot be handled.
Various potential causes have been put forward such as budget cuts; victim-blaming and bias against women and victims of sex crimes; slow workflows due to fear of contamination, and inefficient or redundant testing/auditing requirements; lack of a tracking system for forensic evidence; and misunderstandings of processes to follow. 
In some cases, rape kits have been destroyed before testing without notifying victims, and in some areas of the world, lack of rape kits or trained examiners can also be a problem.
The result can include decreased community trust in police, as well as missed opportunities to identify serial perpetrators and prevent new rapes and other crimes--as rapists are often also guilty of other crimes such as burglary, assault, and murder.

\subsubsection{Health Challenges}
\label{section:health}

Some health problems typically only affect women, such as 
breast cancer, gynecological problems (e.g., premenstrual syndrome (PMS)/premenstrual dysmorphic disorder (PMDD), perimenopause/menopause, dysmenorrhea, endometriosis, ovarian and cervical cancer), and complications due to pregnancy (including difficult childbirth and abortion, as well as perinatal depression)~\cite{barbieri2021gender}.
\footnote{\href{https://online.regiscollege.edu/online-masters-degrees/online-master-science-nursing/womens-health-nurse-practitioner/resources/health-issues-specific-womens-health}{https://online.regiscollege.edu/online-masters-degrees/online-master-science-nursing/womens-health-nurse-practitioner/resources/health-issues-specific-womens-health}}
Other health problems also affect men but disproportionately affect women, such as eating disorders, anxiety, migraines, osteoporosis and autoimmune diseases.

As such, women in the European Union also report poorer health and mental well-being than men, and are more likely to have health limitations over their lifetime~\cite{barbieri2021gender}.
Obtaining a diagnosis also often takes longer for women, possibly due to gender bias.\footnote{\href{weforum.org/agenda/2024/02/womens-health-gap-healthcare}{weforum.org/agenda/2024/02/womens-health-gap-healthcare}}
This might also play some role in why women are more inclined than men to attempt suicide~\cite{baca2010suicidal}.

\subsubsection{Other Day-to-Day Challenges}
\label{section:daily}

Women also deal with inequalities related to the workplace, physical characteristics, and social expectations regarding appearance:
The UN lists eleven largest hurdles for women's equality by 2030, which includes lack of women in leadership positions, workplace discrimination, and an imbalance in unpaid care work~\cite{azcona2023progress}. For example, the report mentions typical rates of 20-30\% of managers being female, women earning globally 51 cents per dollar earned by men, and an estimated extra two hours of unpaid care work each day that women will still be expected to spend by 2050.

Physically, women are also on average smaller, shorter and weaker than men (possessing less muscle mass and bone density and more body fat)~\cite{stevens2010associations}.
This could affect more than day-to-day tasks such as reaching high-up items, carrying heavy objects, and opening tight containers; for example, height also affects perceived dominance, which might explain why most Chief Executive Officers (CEOs) of Fortune 500 companies tend to be tall men~\cite{butera2008height}.

Also regarding appearances, expectations of how women should dress themselves can have repercussions related to crime and health.
Fashion items have been mentioned by criminals seeking to justify their crimes: 
Lacking sufficient pocket space can lead to the use of handbags, which can be seen as easy to snatch and invite theft~\cite{ichikawa2005s}.  
Revealing clothing can also be seen to invite upskirt photography or rape~\cite{peterson2004rape}.
In Sweden, perpetrators sometimes also come from countries with different cultures, where they might never have seen any of their female relatives (e.g., mother or sister) naked or in light clothes, and might never have been heard about norms for dressing in Sweden.
In some countries, women can also be hassled or killed if they do not clothe themselves in a certain way (e.g., by \emph{Mutawa} "religious police" or relatives, if a \emph{hijab} is deemed to not be sufficiently set); 15 girls were reportedly burned alive in 2002 in an incident when the Mutawa beat girls trying to escape from a burning school without their hijabs~\cite{chesler2010ban}.
Also, while tight corsets and footbinding are no longer common, high heels can still lead to reduced control when driving, foot pain, or tripping.

Like with clothing, women are also more targeted by the fashion and cosmetics industries, which are among the biggest international businesses, and encouraged to spend more time on physical appearance (e.g., seeking to avoid tanning in Asia)~\cite{rondilla2007lighter}.
As well, products are sometimes sold at higher prices targeting women that only differ in packaging, fragrance, or color (e.g., razors similar to those sold to men, but just in pink color); such gender-based price segmentation has been described as a "pink tax", and has also been observed to some extent in Sweden~\cite{kardetoft2022pink}.

\subsection{How Robots Could Help Women}
\label{section:howToHelp}

\subsubsection{Robots Helping to Fight Crime}
\label{section:helping_safety}

Robots could interfere with crimes, by increasing women's awareness of criminals and making it more difficult for criminals to escape prosecution (i.e., preventing a perpetrator's ability to carry out an attack and ensuring justice is served).
In regard to voyeurism, hidden spy cameras could be detected, e.g., by checking for unusual objects overlooking sensitive areas (e.g., with reflective lights from lenses, wires, or lights), unknown Wi-Fi devices or radio frequency communications, interference with phone signals, or buzzing sounds.\footnote{\href{https://reolink.com/blog/how-to-detect-hidden-cameras}{https://reolink.com/blog/how-to-detect-hidden-cameras}}
For example, Sami et al. described a smartphone app that uses lasers/time of flight to detect hidden cameras~\cite{sami2021lapd}.
Yu et al. conversely used a thermal camera to detect hidden cameras via heat~\cite{yu2022heatdecam}. 
In the current paper, we also describe a prototype drone aimed at dissuading voyeurism by detecting hidden cameras.
Two differences are that the above approaches require a human to run an app and move around an apartment searching for devices themselves, which might not be easy in hard-to-reach or high locations, and that we also consider detecting object boundaries to reduce the search space.

Likewise, in regard to DV and rape (and possibly also to other crimes such as molestation, stalking, upskirting, and purse-snatching), robots could seek to prevent a woman from being targeted and perpetrators from getting away.
Deterrence could be partially achieved by reducing times in which women at risk must be alone.
For example, a robot capable of recording evidence and calling for help could accompany a woman in risky environments such as a home where DV has occurred, or if she needs to move through a dangerous part of a city, like a parking lot at night.
One option could be to deploy a drone from a nearby rooftop, as police in Southern Sweden are exploring.\footnote{\href{https://www.svt.se/nyheter/lokalt/skane/p-platser-for-polisens-dronare-byggs-pa-hustak-i-malmo}{https://www.svt.se/nyheter/lokalt/skane/p-platser-for-polisens-dronare-byggs-pa-hustak-i-malmo (Swedish)}}
If danger seems likely, a drone could threaten criminals by, e.g., buzzing loudly, flying erratically, moving at head height, and potentially causing injury if a criminal gets too close and causes a collision.
This could be like suddenly being able to summon a loud barking dog to one's aid, with the benefit that a robot could be sent back to its station, or turned off and carried, when the danger is gone.
Additionally, a robot could also indicate a woman's personal space by projecting a circle of light around her, or shine light on a stalker.
One related example in the literature exists, of a "spider dress" designed by Anouk Wipprecht that inflates, extending mechanical arms, based on monitoring proximity and a woman's breathing; another variant releases smoke.\footnote{\href{https://medium.com/@intel/is-that-a-spider-on-your-dress-or-are-you-happy-to-see-me-da25075314b9}{https://medium.com/@intel/is-that-a-spider-on-your-dress-or-are-you-happy-to-see-me-da25075314b9}}

Furthermore, a more complex robot could seek to also protect certain objects, like a woman's drink at a bar; detect if a woman suddenly seems incapacitated and in danger of being abducted; or infer an intent to attack (e.g., hands balled into fists or hidden, getting close, angry/loud language, etc.) 
A remote robot might also be alerted (e.g., by detecting nearby victims' screams~\cite{mathur2022identification}), or be sent by authorities. In such a case, advanced capabilities might be required to distinguish true attacks from jokes, skits, or play-fighting, and make judgements about "distinction" and "proportionality":
A robot might need to identify which person is the victim and which is the perpetrator--a highly challenging task when multiple people are present.
As well, decisions about the level of force to use might include some analysis of force differentials, backdrops and crossfire.
Additionally, age detection could also be used to detect child abuse or forced marriage, and flying robots or soft robots could be useful to enter closed buildings in which trafficking might be occurring.

During a crime, a robot could target the assailant from a difficult angle (e.g., from behind or from above for a drone) with pepper spray or laser to dazzle the assailant's eyes, while emitting loud sounds and bright lights to seek help.
As well, the robot could try to mark the attacker or their vehicle with paint, and record license plates, or interfere with their movement.
The robot could also provide advice during an attack: e.g., to drop to the ground to be harder to move and to avoid being taken to a "second location".

Robots could also help post-hoc, if an attack could not be prevented.
For DV, an AI tool could track hospital records to assess risk, if current laws change. 
Federated learning, e.g., downloading updates to detection models without uploading sensitive or restricted data, could also be one way to ensure that private data are not misused.
After a rape, a robot with a sterile compartment could try to facilitate rape testing immediately, avoiding waiting time at hospitals--possibly with less risk of contamination, since robots lack the DNA of a human investigator. 
As part of this, robots could also visually analyze victims' bodies.
For example, Fernandes et al. reported on a deep learning approach that can identify genital lesions indicative of rape, using a dataset of roughly 400 images collected by the Southern Denmark Sexual Assault Referral Centre (78 from non-consensual and 316 from consensual intercourse)~\cite{fernandes2018deep}.
Robots could also help to search for a hostage or corpse, or track fleeing attackers.
And, in the undesired case that a woman has been murdered, robots are also being built to do forensics on corpses, like the Virtobot system~\cite{ebert2014virtobot}.

\subsubsection{Robots Helping to Support Health}
\label{section:helping_health}

Robots could also help to detect health problems and intervene, in line with the ideas of democratization of health care and data-driven care. 
To detect problems, a robot could generate three-dimensional (3D) scans of a person, e.g., using a Lidar or ultrasound device, along with other sensing.
Scans already possible today using Lidar devices in iPhone Pros or iPads could be facilitated by robots, that can continuously and accurately scan at arbitrary distances, without requiring human time or effort.
Thus, better health outcomes could result by enabling improved methods and continuous measurement:
\begin{itemize}
\item{Outdated, suboptimal methods could be replaced, such as using tape measures to measure belly girth for pregnancy, which can be inaccurate; thus, a robot could perform duties like a midwife, checking a baby's position (upside down or not) in addition to heart rate, etc. As well, AI systems can aid, e.g., in the ultrasound diagnosis of ovarian cancer~\cite{sian16020422artificial}.}
\item{People could measure themselves outside of infrequent doctor visits and have continuous control over their own health data. For example, in some parts of Sweden, patients wait five years between breast X-rays.}
\end{itemize}
In more detail, 3D scans could be used to detect breast cancer, anorexia, and abnormalities:
\begin{itemize}
\item{A lump in a breast indicating potential cancer could be detected by, e.g., a wearable ultrasound scanner.\footnote{\href{news.mit.edu/2023/wearable-ultrasound-scanner-breast-cancer-0728}{news.mit.edu/2023/wearable-ultrasound-scanner-breast-cancer-0728}}}
\item{Weight loss in a short time could indicate anorexia.}
\item{A scan could help women to understand that it's normal to not be shaped like a photo model or porn star (i.e., much variance exists).
For example, this is a goal in the initiative of Visual Sweden called "Visual Vulva".\footnote{\href{https://www.visualsweden.se/en/aktuella-projekt/ar-jag-normal}{https://www.visualsweden.se/en/aktuella-projekt/ar-jag-normal}}}
\end{itemize}
In addition to visual scans, robots have been designed to haptically carry out clinical breast examinations.\footnote{\href{https://www.bristol.ac.uk/news/2023/october/new-robot-could-help-diagnose-breast-cancer-early.html}{https://www.bristol.ac.uk/news/2023/october/new-robot-could-help-diagnose-breast-cancer-early.html}}
As well, for women experiencing morning sickness, a robot could seek to detect causes of ill feeling (e.g., if certain foods are a problem), vomiting, fainting, or overheating.

Where simply scanning is insufficient, robots could intervene in a more complex manner, providing healthcare, positive social interactions and touch, information, or sustenance: 
For example, problems such as ovarian cancer can be tackled with robotic surgery ~\cite{gallotta2023robotic}
Anxiety and depression, which are sometimes exacerbated by loneliness, could be aided by positive social interactions with cuddly robots such as Paro~\cite{joranson2015effects}.
To help women with their periods, a robot could apply heat or massage to minimize cramping, and remind about drinking or offer water to minimize bloating.
To help women with eating disorders, a robot could try to prevent binge eating (e.g., hide food or discourage purchases), provide dietary advice, reinforce a positive body image, stop excessive exercise by positive distraction, or cook with a person.
Cooking could also help women experiencing morning sickness. (Various robots capable of making food exist, including a prototype we developed that checked how salty or sweet its cooking was to ensure healthiness.\footnote{\href {https://www.youtube.com/watch?v=6eqmtkOm-Tk}{https://www.youtube.com/watch?v=6eqmtkOm-Tk}})
As well, a robot could clean after vomiting; catch a fainting woman or call for help; and find places for a woman to rest when walking, like benches in shadow, and fan or cool her.
A childlike embodiment for such a robot could also suggest that a woman is with child even when her bump is not clearly visible.

\subsubsection{Robots Helping with Other Day-to-Day Challenges}
\label{section:helping_daily}

Robots could also support equality by enhancing capabilities and freeing up time:
Exoskeletons or other tools could help to level the playing field, by allowing anyone to be large, tall, and strong.
For example, this might allow smaller women to become accepted as leaders, or peers receiving equal pay, in physically intensive jobs such as construction or policing.
As well, smaller women could reach high-up items, carry heavy loads, and open tight packages--and maybe even better defend against attackers, helping them to fight back and be too heavy to knock down or abduct.
Furthermore, \emph{alloparenting} robots capable of raising children in a good way could also help~\cite{mcclelland2016robotic}--for example, when pay is unequal due to the fear that women will leave work to take care of children, or women are overwhelmed with unpaid care.
For example, a wheeled robot could carry children, while playing with them and providing positive attention.

\subsubsection{Sketches}
\label{section:sketches}

\begin{figure*}
\centering
\includegraphics[width=\textwidth]{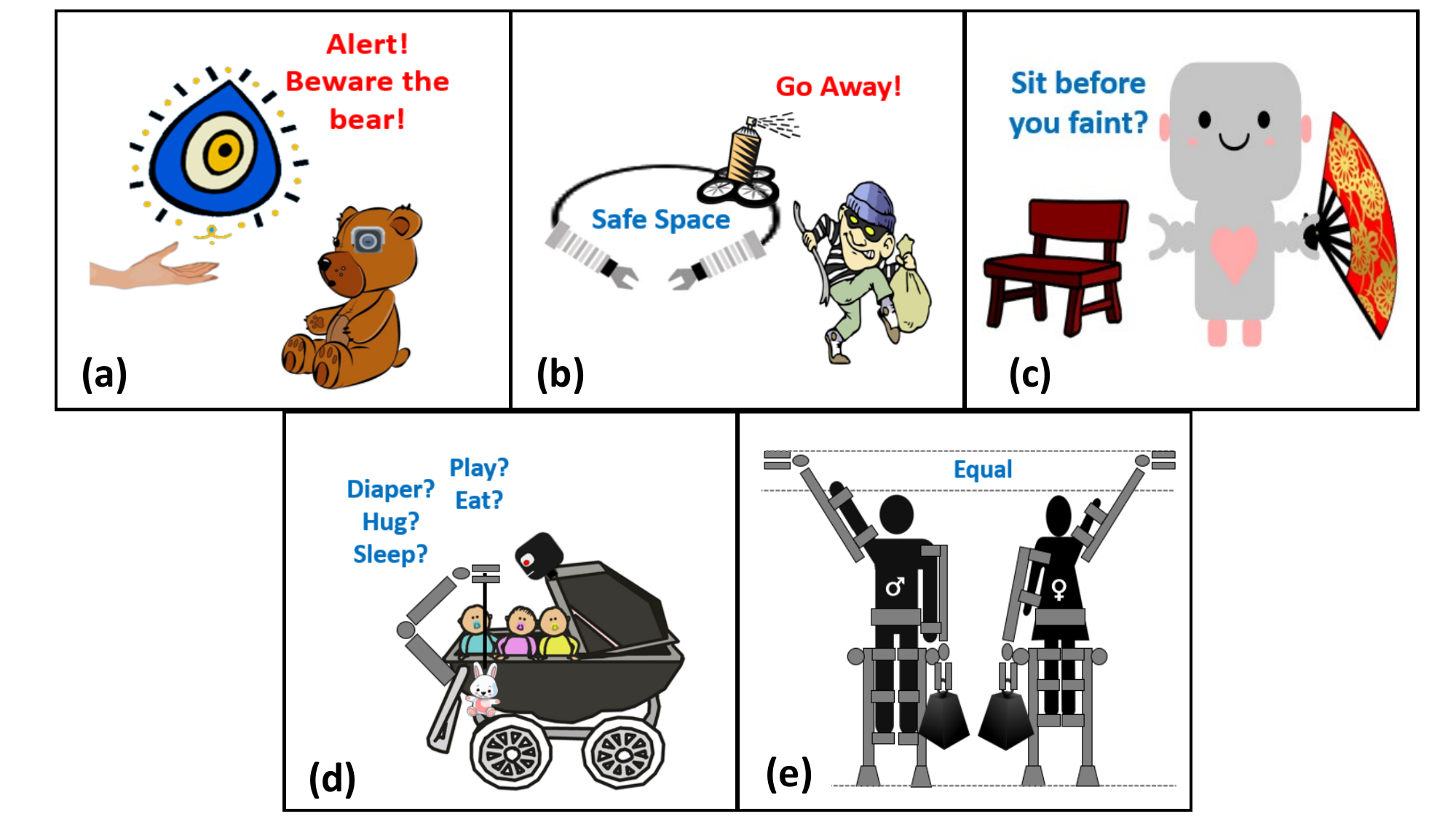}
\caption{Sketches: (a) \emph{Nazar} - A helpful flying "eye" could sense "evil" such as hidden cameras located in high-up or hard-to-see places for women seeking privacy, (b) \emph{Hero} - A robotic barrier could seek to ensure that people intending harm cannot enter a woman's personal space in dangerous places, (c) \emph{Midwife}  - A helper robot could seek to maintain good living conditions for pregnant women, (d) \emph{Allo}  - A robotic baby carriage could help mothers by taking care of repetitive child raising tasks, (e) \emph{Equa-skeleton}  - Exoskeletons could create equal conditions for working women} \label{fig_sketches}
\end{figure*}

From the pool of scenarios, five were selected and transformed into sketches, as shown in Fig.~\ref{fig_sketches}.

\section{PROTOTYPE}
\label{section:prototype}

\begin{figure*}
\centering
\includegraphics[width=.9\textwidth]{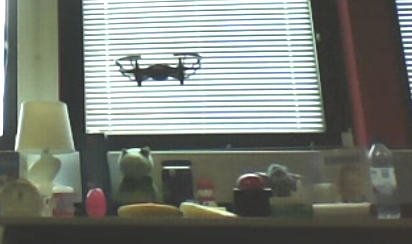}
\caption{Prototype concept: a drone could check a room for hidden cameras} \label{fig_drone}
\end{figure*}

Theoretical ideas alone sometimes miss practical realities that can be exposed by prototyping.
To gain further insight into one of the scenarios, proposed in the first sketch, a proof-of-concept was created, as shown in Fig.~\ref{fig_drone}, in three steps:
\emph{Preparation.} First, a mock-up environment was created containing some objects with hidden cameras and distractor objects intended to make the detection task more difficult.
In total, 22 objects were gathered, a similar number to previous papers:
Toilet Paper Roll, Book, PET Bottle, Pouch (Snowman), Toy Car, Toy Mammoth, Pill Bottle, Toy Frog, Wine Box, Lamp, Toy Box, Clock, Toy Horse, Gum Bottle, Toy Egg, Glasses, Banana, Sponge, Sunscreen, Toy Buzzer, Medicine Box, Paper Roll.  
Some standard Logitech/Plexgear webcams were hidden within five objects: Toilet Paper Roll, Toy Car, Pill Bottle, Wine Box, Lamp.
Furthermore, a simple OpenCV program showing the video feed was run for each camera to simulate recording.\footnote{\href{https://opencv.org}{https://opencv.org}}

\emph{Robot Motion.} Second, we recorded some footage of the drone, a Ryze Tello, flying in front of the objects. We explored various forms of control, e.g., manually controlled the drone via the Ryze Tello smartphone app, as well as using the EasyTello python library\footnote{\href{https://pypi.org/project/easytello}{https://pypi.org/project/easytello}} to issue commands and obtain video, and detecting ArUco markers that could be used to guide the robot or for distance estimation.

\emph{Hidden camera detection.} Third, we explored how spy cameras could be detected using an RGB/thermal camera. 
Since the camera weighed too much (90g without battery vs. 80g payload for the drone), data for our initial exploration were obtained with the camera placed on a desk approximately 1.5m in front of the mock-up environment.

\begin{figure*}
\includegraphics[width=.5\textwidth, height=4cm]{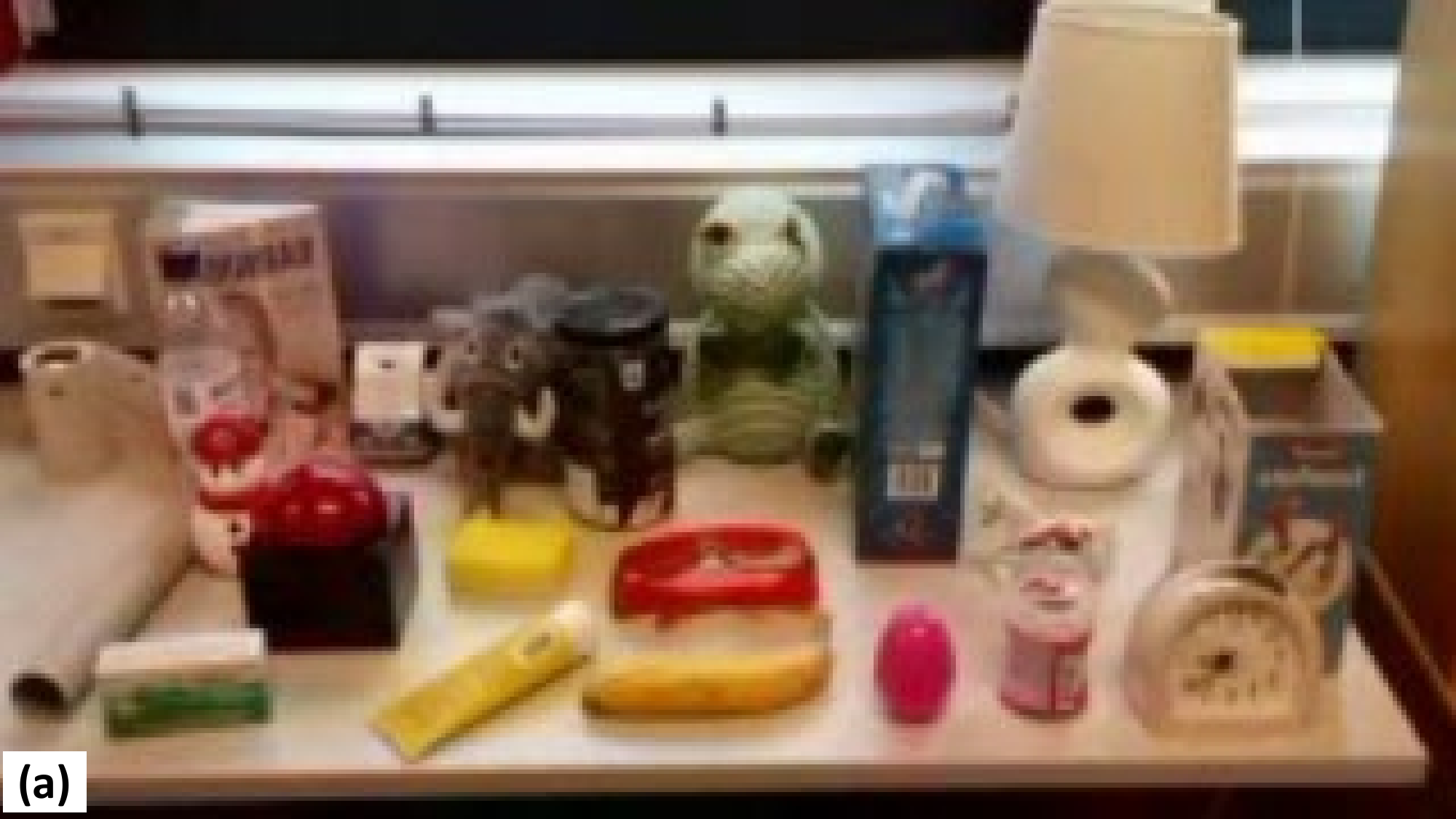}
\includegraphics[width=.5\textwidth, height=4cm]{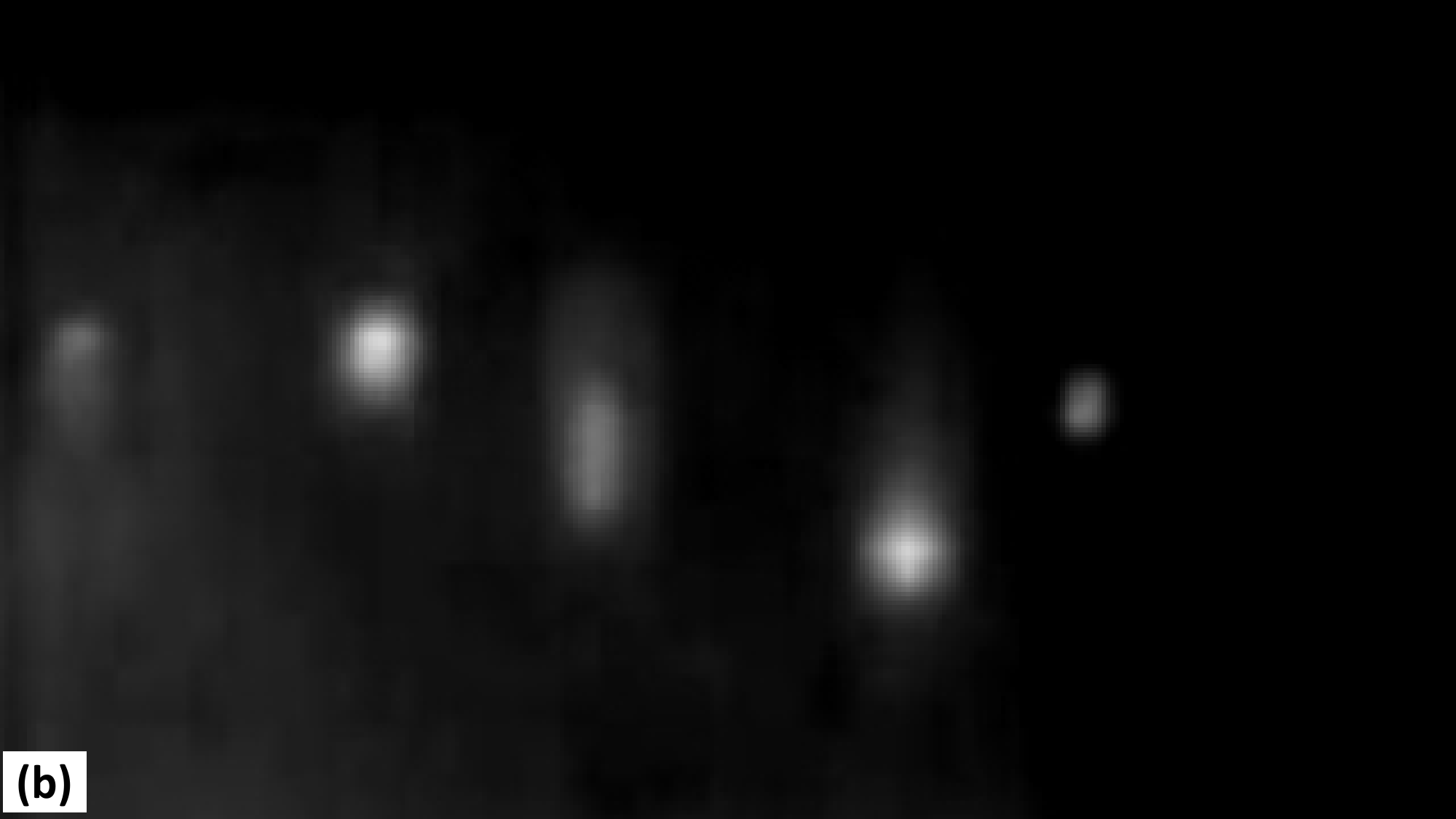}
\includegraphics[width=.5\textwidth, height=4cm]{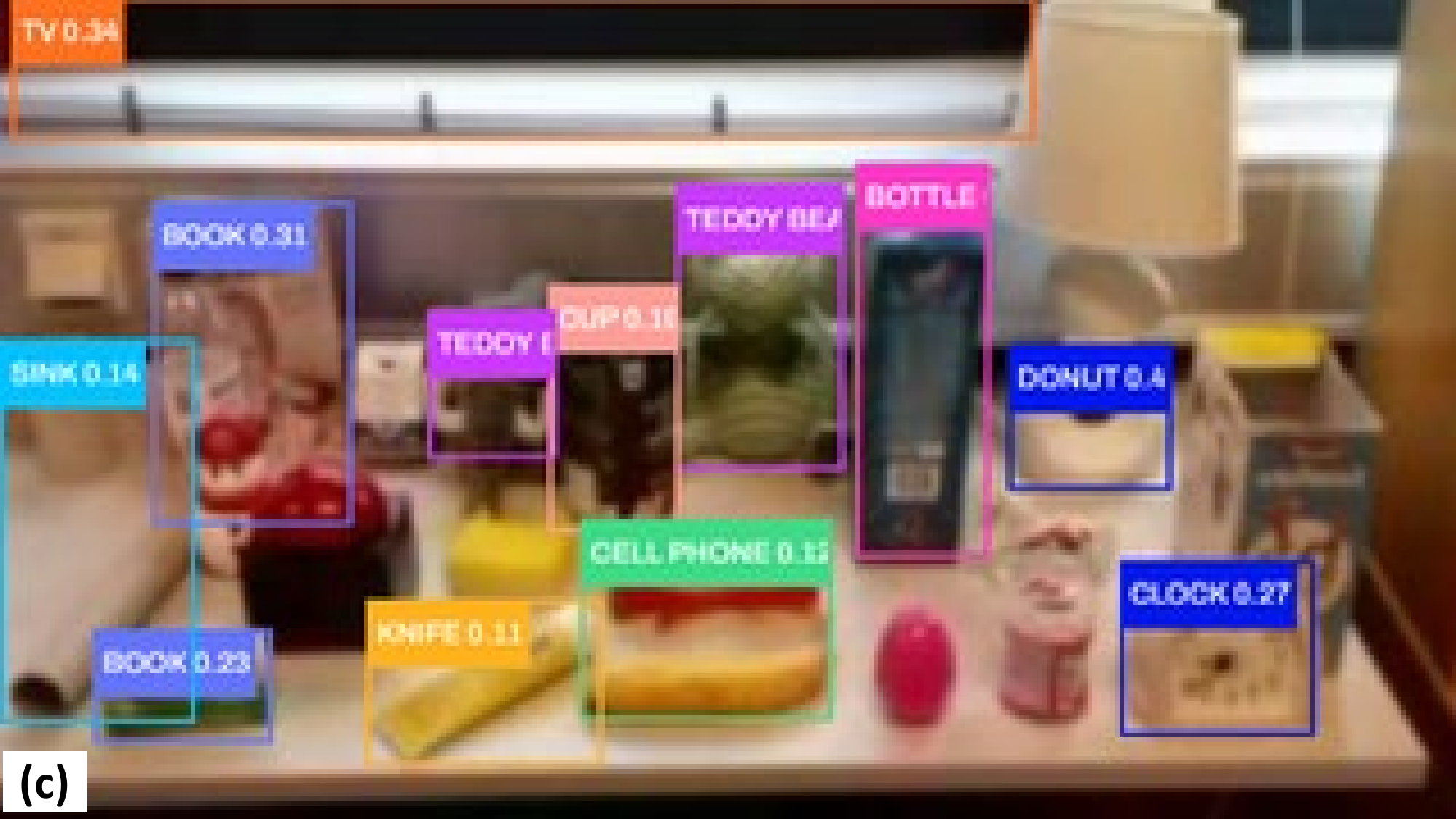}
\includegraphics[width=.5\textwidth, height=4cm]{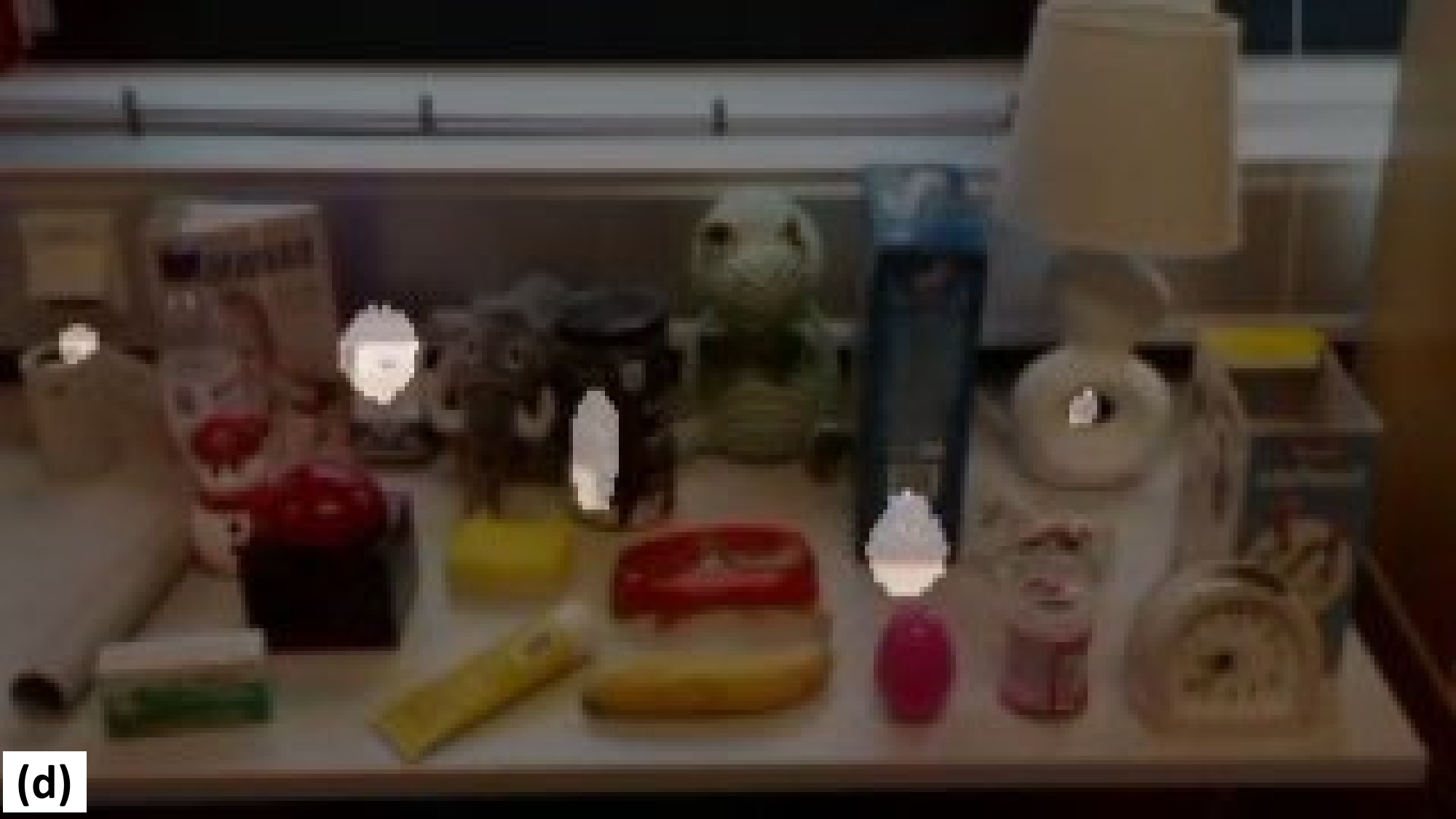}
\includegraphics[width=.5\textwidth, height=4cm]{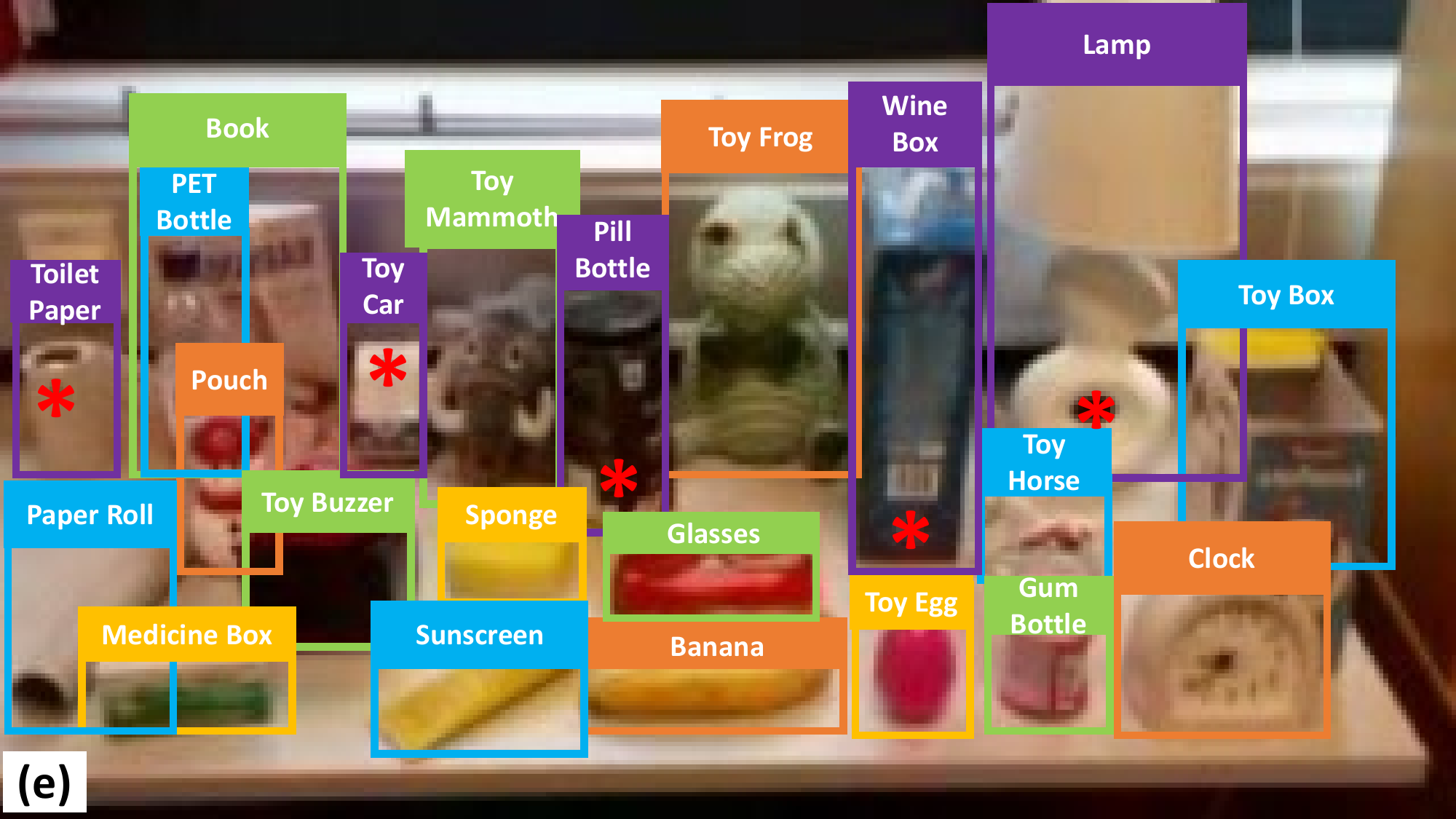}
\includegraphics[width=.5\textwidth, height=4cm]{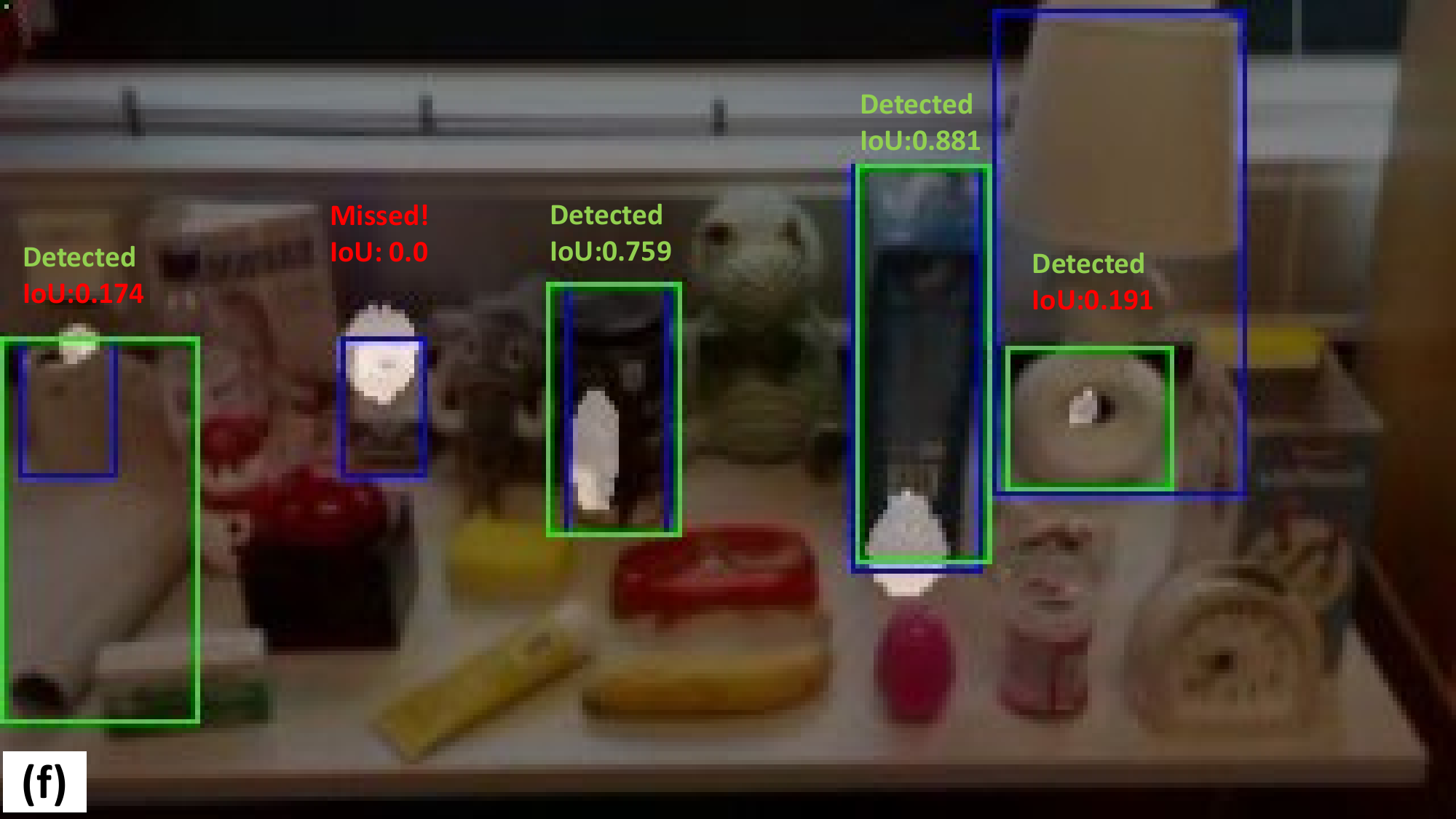}
\caption{Image Processing: (a-b) raw RGB and thermal images, (c-d) intermediate output from YOLO and threshold on thermal traces, (e-f) comparison of overall ground truth on the left--where purple boxes with red asterisks are the target, and distractor objects are labelled in other colors--with the system output on the right in green (ground truth for target objects repeated in blue)} \label{fig_detection}
\end{figure*}

Fig.~\ref{fig_detection} shows this basic process:
Raw RGB and thermal data were obtained using, respectively, a Sony IMX219 8-megapixel sensor, and an inexpensive 80 x 60 FLIR camera capable of detecting heat in the range of 8–14~{\textmu}m.
After calculating a mapping, the thermal camera was used to find warm areas in view that might arise from a hidden camera. 
Simplified thresholding was conducted to derive a mask.
Next, an algorithm detected which objects might be responsible for the warm spots, to reduce the area that needs to be searched and make it easier to find the cameras.
YOLO version 8s from Ultralytics was run on the RGB image with the confidence parameter set to 0.1 to detect the locations of objects as a set of bounding boxes.\footnote{\href{https://www.ultralytics.com/yolo}{https://www.ultralytics.com/yolo}} 
YOLO uses deep learning (a convolutional neural network) to detect objects, where the confidence threshold handles non-maximum suppression.
Finally, bounding boxes enclosing heat traces were selected as the output of the system, representing objects potentially concealing cameras, that a robot could either remove or show to a human.

As a result, 5/5 of the cameras' heat signatures were clearly visible after thermal thresholding, resulting in five detected contours.
Boundary boxes were detected for 4/5 (80\%) of the objects enclosing hidden cameras (although overall, only half of the 22 objects were detected by YOLO).
Finally, the agreement between the ground truth and system output regarding locations of objects concealing hidden cameras was calculated. 
The average Intersection over Union (IoU) was 0.401 (min=0.0 for the Toy Car which was not detected as an object, max=0.881 for the Wine Box).

\section{DISCUSSION}
\label{section:discussion}

The current paper has sought to highlight an important but little-addressed topic: millions of women face problems that could be mitigated via robotics.
Three categories of challenges that disproportionately affect women were identified, in relation to crimes, health, and daily activities--comprising specific challenges such as voyeurism, DV, rape, difficult pregnancies, physical inequalities and unpaid care work.
From these challenges were born ideas of how designers could use robots to better women's lives by interfering with crimes (deterring and bringing to justice), democratizing health care (detecting and intervening), and equalizing opportunities (physically and time-wise).
Five ideas were visualized via concrete sketched examples, illustrating how flying, wheeled, humanoid, and exoskeletal robot designs could aim to help young women seeking privacy and safety, pregnant women, mothers, and working women.
Furthermore, the first sketch was implemented as a prototype, of a drone system that uses a thermal/RGB camera to detect hidden cameras.
A summary of this work is also available via an online video.\footnote{\href{https://youtu.be/7slpfGD1sEU}{https://youtu.be/7slpfGD1sEU}}

\subsection{Reflections on the Prototype}
\label{section:reflectionsOnThePrototype}

Regarding the simplified prototype intended to detect hidden cameras, the result of 80\% accuracy (IoU 0.401) seemed reasonable, given the challenging mock-up environment and simplified detection approach.
We also observed that lowering the confidence parameter allows YOLO to detect more bounding boxes, such that all five objects of interest are detected; a demerit is that this results in many more objects detected, and thereby a higher risk of overlapping bounding boxes and increased complexity.
Furthermore, although the IoU score is imperfect, we believe this should not be a deal-breaker since a robot or person is not limited to checking only the exact inside of each bounding box but can also check the vicinity, if a rough estimate is available of where a camera might be.

Moreover, YOLO's object recognition capability could also be used: 
For example, some alternative heat sources like ovens or people could be removed from consideration; other devices like lamps or computers could be checked for anomalous heat patterns using a "normal" model.
As well, a label could be directly provided to a human in regard to which objects are expected to contain hidden cameras.
However, in our simplified exploration, object recognition appeared to have been challenging, possibly due to illumination, the cluttered mock-up environment, or camera limitations.
As can be seen in Fig.~\ref{fig_detection}, four objects were recognized correctly--two teddy bears, a clock, and a book--whereas seven objects were recognized incorrectly--as a sink, cup, bottle, donut, cellphone, knife, and book. 
This problem could be avoided by instead showing people where cameras might be located (e.g., either using a screen on the robot, or by sending a picture to a person's smartphone).

Along the side, one thing we observed during preparation was that hooking up all five cameras to an old desktop resulted in crashes, possibly due to overwhelming the bandwidth on the same USB bus, so for subsequent attempts several computers were used.
Another observation was that our drone, like other typical drones used by the community, would not able to detect objects on the ceiling due to its camera placements, as it only has two cameras, that look forward and downward respectively.

\subsection{Limitations and Future Work}
\label{section:limitations}

The current work is limited by its exploratory nature:
involving a small group of experts focused on Sweden, and ignoring current practical limitations of robots.
Future studies can gather ideas from a larger pool of female participants with different backgrounds, or automatically identify challenges from the literature using AI methods, and consider factors such as cost, maintenance, battery life, and capabilities.
Moreover, sketches are examples and not "only alternatives": for example, a wheeled robot with long battery life that can hold heavy, powerful sensors could be used in place of a drone if visibility of high-up places is not required.
For the prototype, results are also limited by the controlled lab environment (e.g., detecting expensive, upper-range spy cameras in blurry images from a moving drone might require more complex methods).
Furthermore, the current paper, which falls in the area of HRI and robot design, mostly does not offer technical details of how solutions should be implemented--instead focusing on what we felt was the first fundamental problem, of obtaining a "lay of the land".
(We note too that the paper should not be interpreted as claiming that women's needs should be prioritized over the needs of others; rather the aim is to expose some new challenges whose solutions could benefit all of society.)

In addition to further developing the drone prototype and prototyping the other sketches, future work will explore potential threats and regulation: 
Given that new technologies create not only opportunities, but also potentially new problems, robot designs should factor in ahead of time how robots could fail or be misused; e.g., how to ensure that times or images recorded by a robot are accurate.
Another question is if robots could inadvertently harm the women they seek to help.
For example, wind from a drone's propellers could 
disturb a crime scene, stirring up dust, erasing evidence or introducing contamination.
Or, robots could make it harder for women to escape or hide (e.g., if a costly or large robot crashes, blocks an escape route, or betrays a woman's location).
As well, the concepts in the developed sketches could also be potentially misused to harm women. 
For example, a camera-detecting robot could be used by criminals to test that their cameras are well-hidden and will not be easily detected.
Or, a robot that guards a woman's personal space could be hacked to slip a sedative into her drink.
Furthermore, criminals could also potentially misuse robots to get others in trouble (e.g., using disguises to trick a robot into thinking an innocent person committed a crime).
Thus, one important area of future work will involve refining such designs, such that, e.g., robots could facilitate legal or healthcare processes by delivering new or better forms of evidence in courts or hospitals.

Another central question regards who will control women's ability to protect themselves with robots.
Although governments introduce various rules and regulations, these might not "set the pace", since criminals can ignore rules, and people often use apps, which have their own terms decided by technology providers.
While users cannot be aware of all sets of terms for the apps they use, even if terms might be in conflict, it seems like women themselves could set up the apps and robots they need for protection, e.g., detecting rapes in real-time with a drone. And, if such usage becomes widespread, it could motivate modifying existing restrictions.
Therefore, we believe that prototyping solutions in this area could be a useful showcase for what AI can do to help people.

In summary, we believe that further exploration of how technologies can be used to help various groups, including women, will contribute to realizing a safer, healthier, more inclusive, and better future for all of society.

\addtolength{\textheight}{-0cm}   







\bibliographystyle{IEEEtran}
\bibliography{IEEEabrv,robots_for_women}





\end{document}